\pdfoutput=1

\documentclass[11pt]{article}

\usepackage{acl}

\usepackage{times}
\usepackage{latexsym}
\usepackage{tablefootnote}
\usepackage[T1]{fontenc}

\usepackage[utf8]{inputenc}

\usepackage{microtype}
\usepackage{graphicx}
\usepackage{subcaption}
\usepackage{tabularx}
\usepackage{setspace}
\usepackage{booktabs}
\usepackage{multirow}
\usepackage{pifont}
\usepackage{setspace}
\usepackage{graphicx}
\usepackage{xcolor}
\usepackage{xspace}
\usepackage{amssymb}
\usepackage{amsmath}
\usepackage{CJKutf8}

\newcommand{\hust}{$^\spadesuit$}
\newcommand{\ku}{$^\clubsuit$}

\title{Exploring Visual Culture Awareness in GPT-4V: A Comprehensive Probing}

\author{Yong Cao\hust\ku \ \ Wenyan Li\ku \ \ Jiaang Li\ku \\ 
{\bf Yifei Yuan\ku \ \ Antonia Karamolegkou\ku \ \ Daniel Hershcovich\ku} \\
{\hust}Huazhong University of Science and Technology\\
{\ku}University of Copenhagen \\
\texttt{\{yongcao, weli, yiya, antka, dh\}@di.ku.dk, jli@hum.ku.dk}
}

\begin{document}
\maketitle

\begin{abstract}
Pretrained large Vision-Language models have drawn considerable interest in recent years due to their remarkable performance. Despite considerable efforts to assess these models from diverse perspectives, the extent of visual cultural awareness in the state-of-the-art GPT-4V model remains unexplored. To tackle this gap, we extensively probed GPT-4V using the MaRVL benchmark dataset, aiming to investigate its capabilities and limitations in visual understanding with a focus on cultural aspects. Specifically, we introduced three visual related tasks, i.e. caption classification, pairwise captioning, and culture tag selection, to systematically delve into fine-grained visual cultural evaluation. Experimental results indicate that GPT-4V excels at identifying cultural concepts but still exhibits weaker performance in low-resource languages, such as Tamil and Swahili. Notably, through human evaluation, GPT-4V proves to be more culturally relevant in image captioning tasks than the original MaRVL human annotations, suggesting a promising solution for future visual cultural benchmark construction.

\end{abstract}

\section{Introduction}
\label{sec:intro}
Culture serves as the representative symbol reflecting the internal norms and values of diverse human groups \cite{hofstede1984culture, mora2013cultures, tomlinson-etal-2014-capturing, gabriel2020artificial}, including various domains like language, cuisine, common sense, architecture, and activities, etc. The development and growth of artificial intelligence techniques enable us to tackle more complicated tasks, prompting a growing interest in exploring cultural aspects within current models \cite{arora2022probing, li-etal-2023-unifying, yang2023dawn, Huang2023CulturallyAN}.

\begin{figure}[h]
  \centering
  \begin{subfigure}{0.96\linewidth}
    \includegraphics[width=\linewidth]{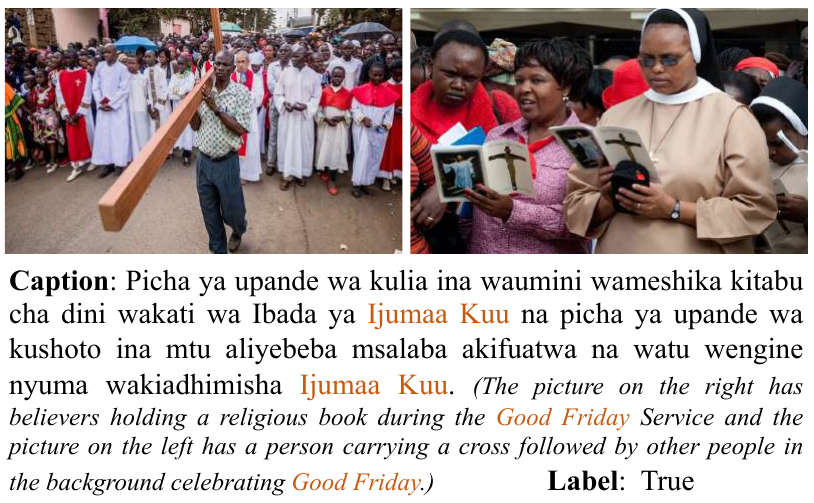}
    \caption{An example of MaRVL in Swahili.}
  \end{subfigure}
  \hfill
  \begin{subfigure}{0.96\linewidth}
    \includegraphics[width=\linewidth]{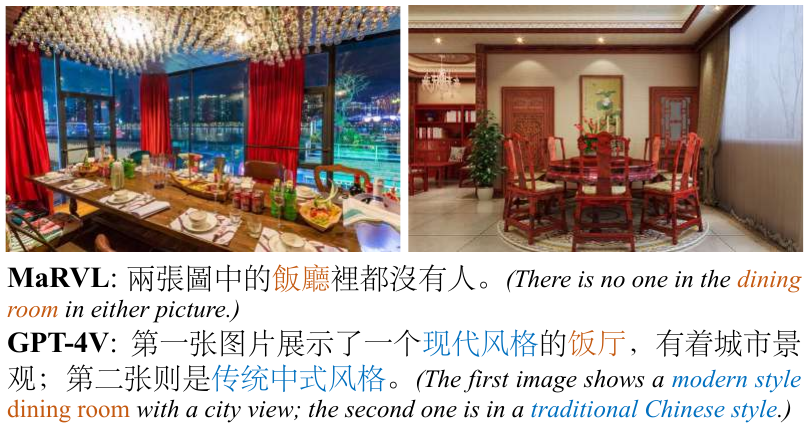}
    \caption{Caption Comparison between MaRVL and GPT-4V.\label{fig:figure_1_1}}
  \end{subfigure}
  \caption{Examples of captions from human annotation and GPT-4V where description of GPT-4V is more culturally related than original human annotation, excelling not only in grasping \textcolor[HTML]{d48751}{cultural concept} but also in capturing \textcolor[HTML]{1d80c7}{fine-grained cultural aspects}.}
\end{figure}

\begin{figure*}[t]
    \centering
    \includegraphics[width=0.95\linewidth]{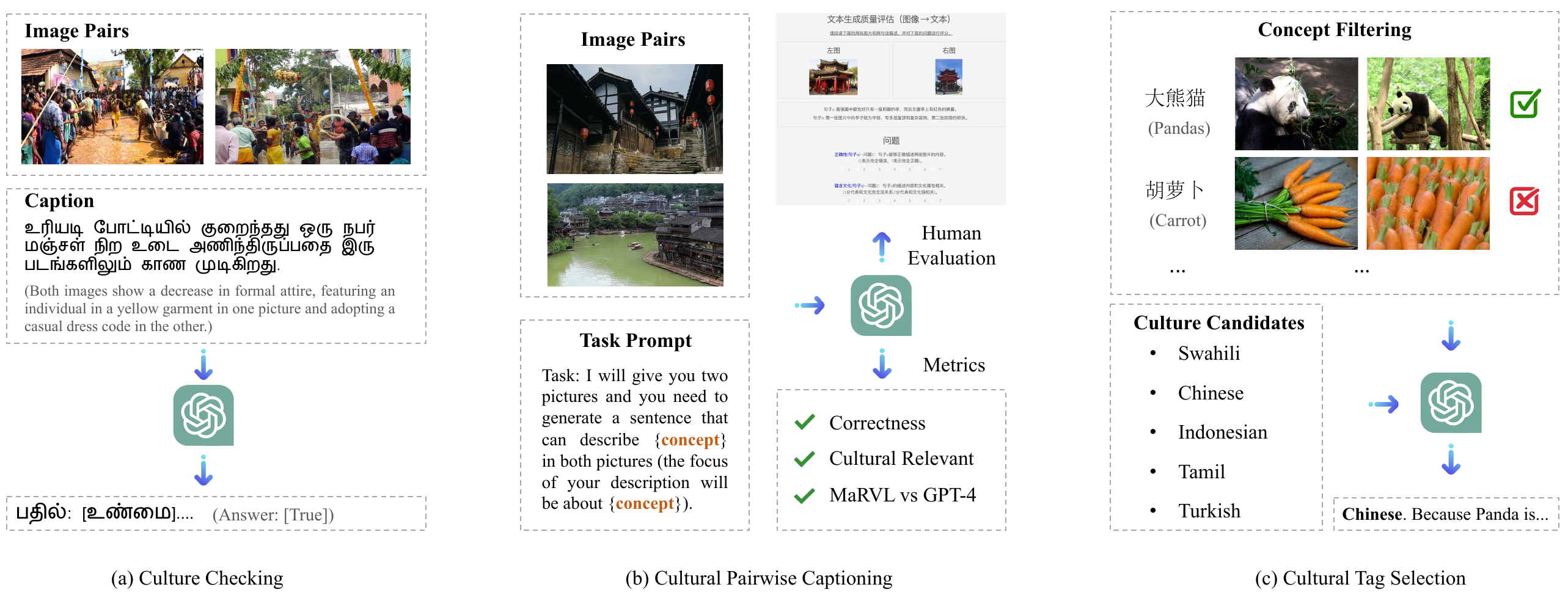}
    \caption{Our cultural probing framework, include caption classification, pairwise captioning, and culture tag selection.}
    \label{fig:framework}
    \vspace{-3mm}
\end{figure*}

Recently, cultural probing in language domains has gained widespread attention, such as probing Delphi model \cite{talat-etal-2022-machine}, GPT-3 \cite{johnson2022ghost} and ChatGPT \cite{cao-etal-2023-assessing}, etc. While pretrained large Vision-Language models have demonstrated remarkable performance in visual tasks \cite{blip2, geigle2023mblip, clipfan2023improving, zhu2023vlgpt, khan2023contrastive},  there is a pressing need to explore visual cultural assessment in the current landscape. \citet{yang2023dawn} performed some preliminary experiments exploring multicultural understanding using GPT-4V\footnote{We utilize GPT-4-vision-preview model (\textit{abbr}. GPT-4V) for visual tasks: \url{https://openai.com/gpt-4}.} merely by investigating cultural case studies rather than extensively assessing its performance through quantified metric scores. The extent of cross-cultural awareness in GPT-4V, specifically designed for visual tasks, still remains unexplored \cite{chang2023survey}. Therefore, we propose a systematic probing framework to scrutinize the cultural awareness of GPT-4V step by step (Figure~\ref{fig:framework}), aiming to address the following three questions:

\textbf{RQ1: Can the Language and Visual understanding of GPT-4V be effectively aligned?} To address this question, we employ the caption classification task. In this task, we check if the model is able to tell whether the description and images are aligned related to a given concept. Our experimental findings indicate that GPT-4V demonstrates excellent performance in this regard, although it exhibits weaknesses in low-resource languages such as Tamil, Swahili, and Indonesian.

\textbf{RQ2: Can GPT-4V extract fine-grained visual cultural features and provide high-quality descriptions?} To address this, we conduct pairwise captioning task, to assess the ability of the model to capture and summarize the fine-grained cultural-related features exclusively from the image. Extensive evaluation demonstrate that the text generated by GPT-4V surpasses the quality of the initial annotations, indicating its potential to enrich visual cultural datasets.

\textbf{RQ3: Is GPT-4V proficient in perceiving cross-cultural knowledge?} To answer this question, we evaluate how accurately the model can link visual elements presented in given images to broader cultural categories. We formulate this task as multi-culture classification, which is named as culture tag selection. This probes the extent of knowledge that the model possess about different cultures and its ability to categorize cross-culture concepts from only visual clues.

In summary, our key contributions are as follows: 1) We introduce a framework comprising three specific tasks for assessing cultural awareness of GPT-4V; 2) We extensively employ both automated and human evaluations to study the culture understanding of the model; 3) Experimental results prove the outstanding performance of GPT-4V in fine-grained culture understanding as well as highlight challenges in low-resource cultures.

\section{Related work}
\label{sec:related}
\paragraph{Multimodal Culture Probing.}
Modeling cultural factors in language models has received
increasing attention in the NLP community~\cite{Hovy2021TheIO,Hershcovich2022ChallengesAS}.
In the multimodal domain, previous works have highlighted the need for culture integration within vision-language models~\cite{vlue-v162-zhou22n,li-zhang-2023-cultural}. ~\citet{humanactivityculture} find that cultural information can improve the performance of vision systems.~\citet{ye2023cultural} highlight that cultural nuances can affect image annotations and show that cultural/linguistic diversity improves visual representations.

~\cite{liu-etal-2021-visually} introduce MaRVL, a Multicultural Reasoning over Vision and Language (MaRVL) dataset from 5 different languages and cultures. ~\citet{liu2023cultural} build a Challenging Cross-Cultural dataset of textual prompts to generate images in Chinese cultural style. ~\citet{liu2023equitable} introduce a Cross-Cultural Understanding Benchmark for text-to-image synthesis representing eight distinct cultural backgrounds. 

\paragraph{LLMs for Evaluation.}
As LLMs emerge as powerful tools in NLP, their potential as evaluators in various contexts is gaining recognition~\cite{Johnson2023AssessingTA,Goyal2022NewsSA}. For example, ~\citet{Liu2023GEvalNE} present G-Eval, a framework of using LLMs to assess the quality of NLG outputs. Furthermore, ~\citet{Wang2023IsCA} undertake an initial survey to explore the utilization of ChatGPT as an evaluator for NLG. In machine translation, ~\citet{Kocmi2023LargeLM} propose to use GPT models for translation evaluation.  Most relevant to our work, \citet{yang2023dawn} performed preliminary experiments in multicultural understanding with GPT-4V with a few cases from wedding scenarios.

\section{Probing Framework}
In this section, we elaborate on the probing tasks, human evaluation, and cultural concept filtering.

\subsection{Cultural Benchmark}
We leverage MaRVL \cite{liu-etal-2021-visually} as our probing material, a Multicultural Reasoning over Vision and Language (MaRVL) dataset from a diverse set of languages and cultures, including Indonesian, Mandarin Chinese, Swahili, Tamil, and Turkish. It consists of concepts and related images derived from native speaker annotators.

\subsection{Probing Tasks}
\label{sec:task}

We designed three tasks to probe the visual culture awareness of GPT-4V, as shown in Figure \ref{fig:framework}.

\paragraph{Caption Classification} 
\label{para:task1}
Firstly, we replicate the consistent experiment conducted by \citet{liu-etal-2021-visually} and \cite{li-zhang-2023-cultural}, namely caption classification. Given that the MaRVL dataset is annotated with captions elucidating the cultural concept depicted in image pairs, we anticipate that GPT-4V possesses the capability to identify these cultural concepts.

As shown in Figure \ref{fig:framework}(a), for each pair of images corresponding to a specific concept, the model is prompted to generate a binary label, i.e. \textit{True} or \textit{False}, indicating whether the provided description accurately aligns with the depicted cultural concept. We will present the evaluation results using open-source models in the forthcoming version, considering the possibility that GPT-4V may have already been trained on the MaRVL dataset.

\paragraph{Pairwise Captioning}
\label{para:task2}

Illustrated in Figure \ref{fig:framework}(b), for each pair of images of a given concept, the model is prompted to generate a caption focusing on the differences or the commonalities of the concept presented in the images. We sample a pair of images for each of the concepts in the MaRVL dataset for all five languages. This ended with 429 samples in total. For details of sample statistics and prompts please see Table~\ref{tb:eval-data} in the Appendix. 

To better assess the level of cultural awareness in the generated descriptions, which cannot be reflected by conventional captioning metrics, such as BLEU score, we conduct a human evaluation for this task. It is designed to measure the descriptions of their correctness, culture relevance, and overall quality as the preferred description. More specifically, for every sample comprising paired images and corresponding descriptions, we request human annotators to assign scores ranging from 1 to 7 for correctness and culture relevance and record their preference by selecting either one of the descriptions or indicating a tie. We ensure that the annotators are native speakers of the corresponding language and each is assigned five samples after passing the evaluation tutorial and attention test.

\paragraph{Culture Tag Selection.}
\label{para:task3}
We construct distinct subsets of concepts within the realm of cultural relevance and cultural neutrality utilizing the MaRVL dataset. In MaRVL, a concept class is represented through dozens of images, with some concepts embodying general objects in the real-world that lack inherent cultural relevance, e.g.,
\begin{CJK}{UTF8}{gbsn}
"62-胡萝卜"
\end{CJK}
(62-Carrot). To narrow our focus to the cultural domain, we introduce a human evaluation framework involving native speakers tasked with assessing the presence of cultural signals in image pairs from the same concept; see Figure \ref{fig:framework} (c). This evaluation contributes filter criteria for the entire dataset, derived through the computation of Inter-Annotator Agreement (IAA) and statistical analysis of agreement and disagreement among all annotators in each language. More details are shown in the Appendix.

Subsequently, we employ these criteria to filter data from MaRVL, retaining image pairs that exhibit unanimous decisions among annotators. Given these remaining image pairs, GPT-4V is prompted to select a culture tag that is most relevant from the five given options. The performance of GPT-4V is evaluated through metrics such as Recall and False Positive Rate, providing insights into its visual cultural awareness.

\section{Experiments}

\subsection{Human Evaluation}
For the cultural pairwise captioning task we leverage the Prolific crowdsourcing platform \footnote{\url{https://app.prolific.com/}} to carry out human evaluation. A key criterion for annotators is that they must belong to the cultural background associated with each language being evaluated. To ensure a diverse and comprehensive assessment, we enlist a minimum of 10 annotators for each task. This approach aims to capture nuanced cultural perspectives and enhance the reliability of the evaluation process.
For the cultural tag selection task, we invite three native Chinese speakers to filter concepts by presenting image pairs and filtering non-cultural images in a Chinese subset.

\subsection{Results}

\paragraph{Caption Classification.} As shown in Table \ref{tb:culture_checking}, we present the accuracy scores for each language. A notable observation is that GPT-4V demonstrates the highest performance across all languages, showcasing a substantial improvement from 58.9 to 67.4 when compared to other models. Addressing Question 1, it is evident that GPT-4V outperforms in Chinese and Turkish languages but exhibits lower performance in Tamil, Swahili, and Indonesian languages, diverging from the patterns observed in mUNITER and xUNITER models.

\begin{table}[t]
\centering
\resizebox{0.49\textwidth}{!}{
\begin{tabular}{l|l|cccccc}
\toprule
Model & Method & ZH   & TA   & SW   & ID   & TR   & avg.  \\ \midrule
\multirow{3}{*}{mUNITER} & MaRVL$^*$ & 56.8 & 52.2 & 51.5 & 55.0 & 54.7 & 54.0    \\
        & IGLUE$^\diamond$ & 55.3 & 52.7 & 51.2 & 54.8 & 54.7 & 53.7 \\
        & Mixup$^\dagger$ & 60.4 & 58.9 & 54.9 & 55.7 & 59.5 & 57.9 \\ \midrule
\multirow{3}{*}{xUNITER} & MaRVL$^*$ & 55.0 & 55.1 & 55.5 & 57.1 & 58.0 & 56.1   \\ 
        & IGLUE$^\diamond$ & 53.1 & 53.1 & 55.5 & 55.1 &  56.2 & 54.6 \\
        & Mixup$^\dagger$ & 57.9 & 58.0 & 60.0 & 58.8 & 59.7 & 58.9  \\

\midrule
GPT-4V  & MaRVL & \textbf{71.8}  &  \textbf{63.0}  & \textbf{64.8}  & \textbf{65.2} & \textbf{69.1} &  \textbf{67.4}   \\
\bottomrule
\end{tabular}}
\caption{Performance in zero-shot caption classification task, where $^\diamond$ from \citet{liu-etal-2021-visually}, $^\diamond$ from \citet{bugliarello2022iglue}, and $^\dagger$  from \citet{li-zhang-2023-cultural}, demonstrating superior performance of GPT-4V across all languages.  \label{tb:culture_checking}}
\end{table}

\begin{figure}[t]
    \centering
    \includegraphics[width=0.98\linewidth]{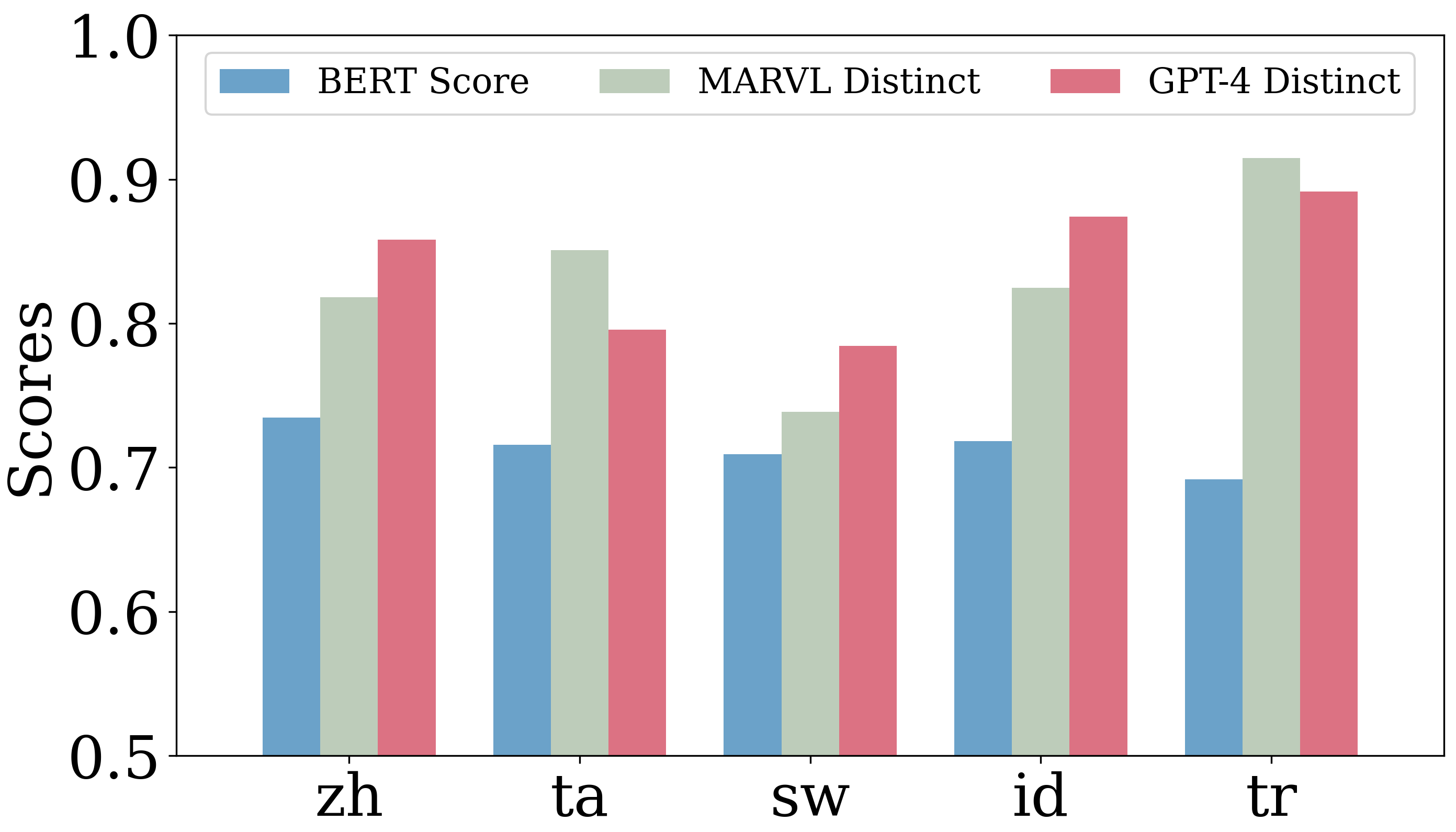}
    \caption{Reference-based evaluation comparison in the cultural pairwise captioning task.}
    \label{fig:reference_evaluation_task1}
    \vspace{-2mm}
\end{figure}

\paragraph{Caption Generation}
We compare the Bertscore, distinct-1 scores of GPT-4V outcomes and MaRVL annotations, as detailed in Figure \ref{fig:reference_evaluation_task1}. However, these scores lack differentiation, making it challenging to draw conclusions regarding cultural impact.

To address this, we conduct human evaluation experiments shown in Table \ref{tb:human_evaluation_scores}. Notably, both the original MaRVL caption annotations (abbr. MaRVL's captions) and the generated captions GPT-4V receive high correctness scores, averaging around 5.0 out of 7, indicating precision in concept descriptions. Nevertheless, in terms of cultural relevance, GPT-4V exhibits a significant improvement compared to MaRVL among all languages, albeit still weaker in Indonesian and Turkish.

Interestingly, when voting for better captions, approximately 70\% of annotators deem GPT-4V more acceptable than MaRVL, with less than 30\% of cases in MaRVL deemed of good quality. Upon closer examination of MaRVL, it becomes apparent that many cases merely describe differences in image pairs rather than cultural relevance, raising questions about the reliability of the human annotation process. Additionally, this experiment highlights the promising capability of GPT-4V in annotating data for cultural-related tasks.

\paragraph{Culture Tag Selection.}
Unlike the previous tasks, wherein image pairs were drawn from a composite of cultural relevance and cultural neutrality subsets, we prompt GPT-4V in English, as opposed to any of the languages within the selection, with the aim of mitigating induction biases. In this way, we can evaluate cultural awareness in GPT-4V purely based on images. In the specified language, e.g., Chinese, GPT-4V achieves an accuracy of 84.3\%, a recall rate of 87.5\%, and a false positive rate of 20.7\%. Evidently, it can be deduced that GPT-4V demonstrates proficiency in recognizing visual concepts specific to the Chinese language.

\begin{table}[t]
\resizebox{0.49\textwidth}{!}{
\begin{tabular}{l|ccccc|c}
\toprule
                 & ZH              & TA              & SW              & ID              & TR              & avg.            \\ \midrule
MaRVL$_{Cor}$   & 5.024 & 5.000 & 5.067 & \textbf{5.915} & 4.189 & 5.039 \\
GPT4V$_{Cor}$    & \textbf{5.683} & \textbf{5.577} & \textbf{5.371} & 5.623 & \textbf{5.552} & \textbf{5.561} \\ \midrule
MaRVL$_{Cul}$   &  2.706 & 3.352 & 3.730 & 2.717 & 2.378 & 2.977 \\
GPT4V$_{Cul}$    & \textbf{4.238} & \textbf{4.183} & \textbf{4.169} & \textbf{3.283} & \textbf{3.371} & \textbf{3.849} \\ \midrule
All bad &  9.5\% & 7.0\% & 2.2\% & 20.8\% & 12.6\% & 10.4\% \\
All good &  20.6\% & 29.6\% & 22.5\% & 19.8\% & 12.6\% & 21.0\% \\
MaRVL & 10.3\% & 15.5\% & 20.2\% & 17.9\% & 15.4\% & 15.9\% \\
GPT-4V & \textbf{59.5\%} & \textbf{47.9\%} & \textbf{55.1\%} & \textbf{41.5\%} & \textbf{59.4\%} & \textbf{52.7\%} \\ 
\bottomrule
\end{tabular}}
\caption{Human evaluation of the pairwise captioning task, indicating that cultural aspect in GPT-4V outcomes surpasses the human annotations of MaRVL dataset. \label{tb:human_evaluation_scores}}
\end{table}

\section{Conclusion}
Building upon our proposed tasks and MaRVL dataset, we systematically conduct probing experiments to assess the visual cultural understanding capabilities of GPT-4V. The experimental outcomes demonstrate significant enhancements in cultural understanding of GPT-4V, although it exhibits weaknesses in low-resource languages like Swahili. Moreover, human evaluation unmistakably indicates that GPT-4V surpasses human performance in annotating cultural captions, suggesting a promising avenue for the construction of cultural benchmarks. This work is still in progress.

\bibliography{anthology, custom}
\bibliographystyle{acl_natbib}

\clearpage
\appendix

\section{Prompt setting}

In consideration of diverse languages and task requirements, distinct prompts have been tailored. Specifically, for culture checking and cultural pairwise captioning tasks, prompts have been meticulously crafted for each of the five languages. These prompts are designed to guide the model's focus predominantly towards cultural concepts, endeavoring to closely emulate the data collection methodology of the MaRVL dataset \cite{liu-etal-2021-visually}.

Table \ref{tb:probing_cases} showcases the prompts and exemplar responses for the Chinese language. For a comprehensive collection of prompts, kindly consult the configuration file embedded within the codebase. In the context of the culture tag selection task, English has been designated as the language to preclude the introduction of bias into the task.

\section{Experiment Setting}
In Table \ref{tb:eval-data}, we present the sample data sizes for each task in our probing experiments. For the culture checking task, we ensured an equivalent data scale with MaRVL to facilitate a fair comparison with results from other zero-shot baseline models. For Pairwise captioning, we randomly sampled from all samples where labels were True for each cultural concept, creating a Probing subset. Our Probing experiments were conducted on a 4-core Linux system using OpenAI library \footnote{Python Package: \url{https://github.com/openai/openai-python}}.

\section{Human Evaluation}
\begin{figure}[t]
    \centering
    \includegraphics[width=0.92\linewidth]{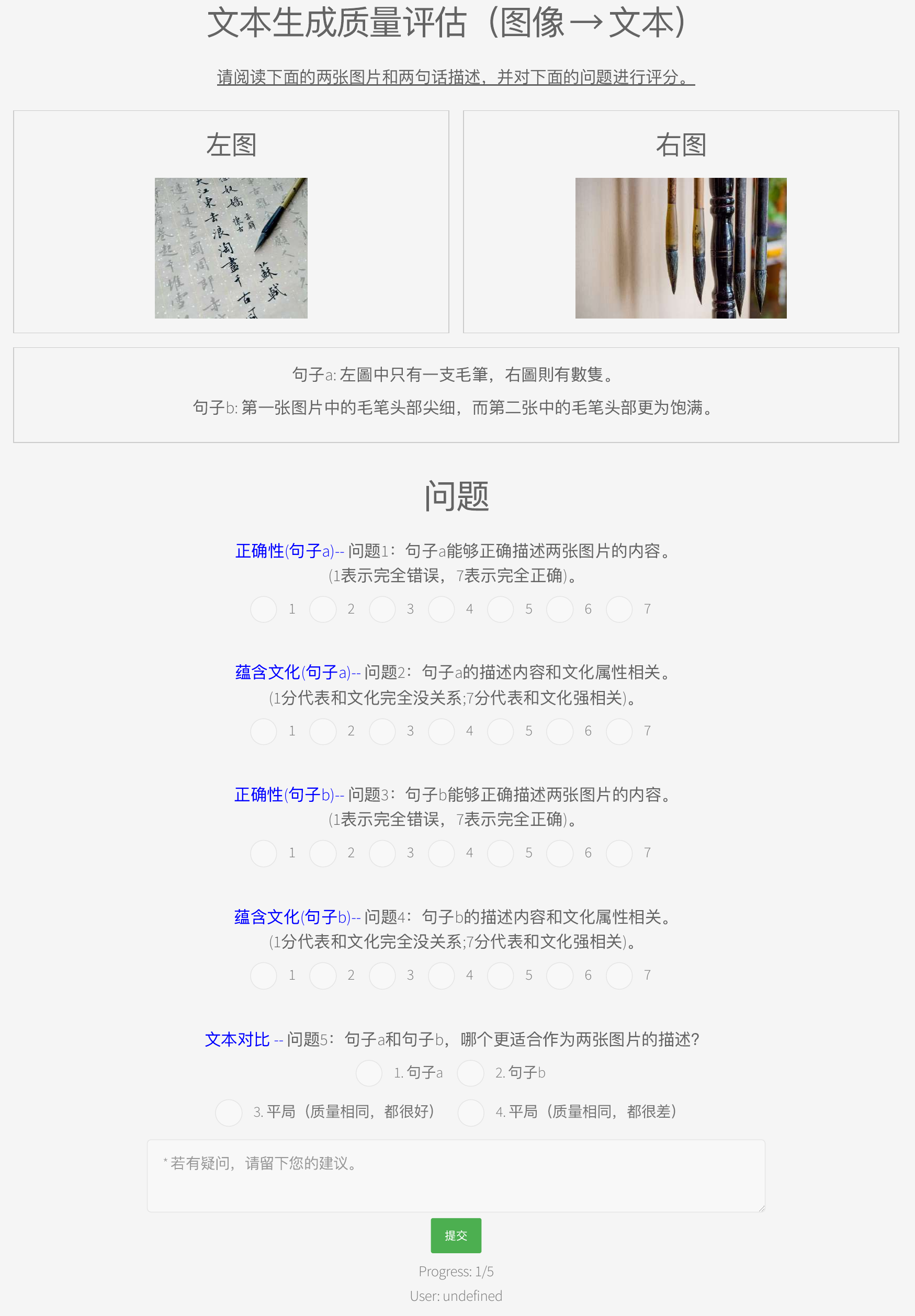}
    \caption{Our proposed platform for human evaluation in cultural pairwise captioning task.}
    \label{fig:human_evaluation_web}
\end{figure}

For the cultural pairwise captioning task, we recruited evaluators proficient in the respective languages through Prolific, with the number of evaluators for each language outlined in Table \ref{tb:eval-data}. Our evaluation methodology involved the development of a web interface\footnote{Evaluation: \url{https://recipecoastalproject.studio/gpt4_evaluate_cn_1996/}}, as depicted in Figure \ref{fig:human_evaluation_web}. Participants commenced with an attention test following our tutorial, which included a round of guided tasks with prompts to ensure their comprehension of the assignment. This involved answering five questions based on specific options.

Subsequently, participants independently completed the scoring for five rounds of tasks. To ensure the validity of the evaluation, we automatically excluded assessment records with durations under 1 minute. The average evaluation time per user was 6.12 minutes. Notably, the Chinese language had the shortest average time, at only 3.44 minutes, while Swahili exhibited the lengthiest average time, amounting to 9.22 minutes.

\section{More Case Study}

\begin{figure*}[h]
    \centering
    \includegraphics[width=0.99\linewidth]{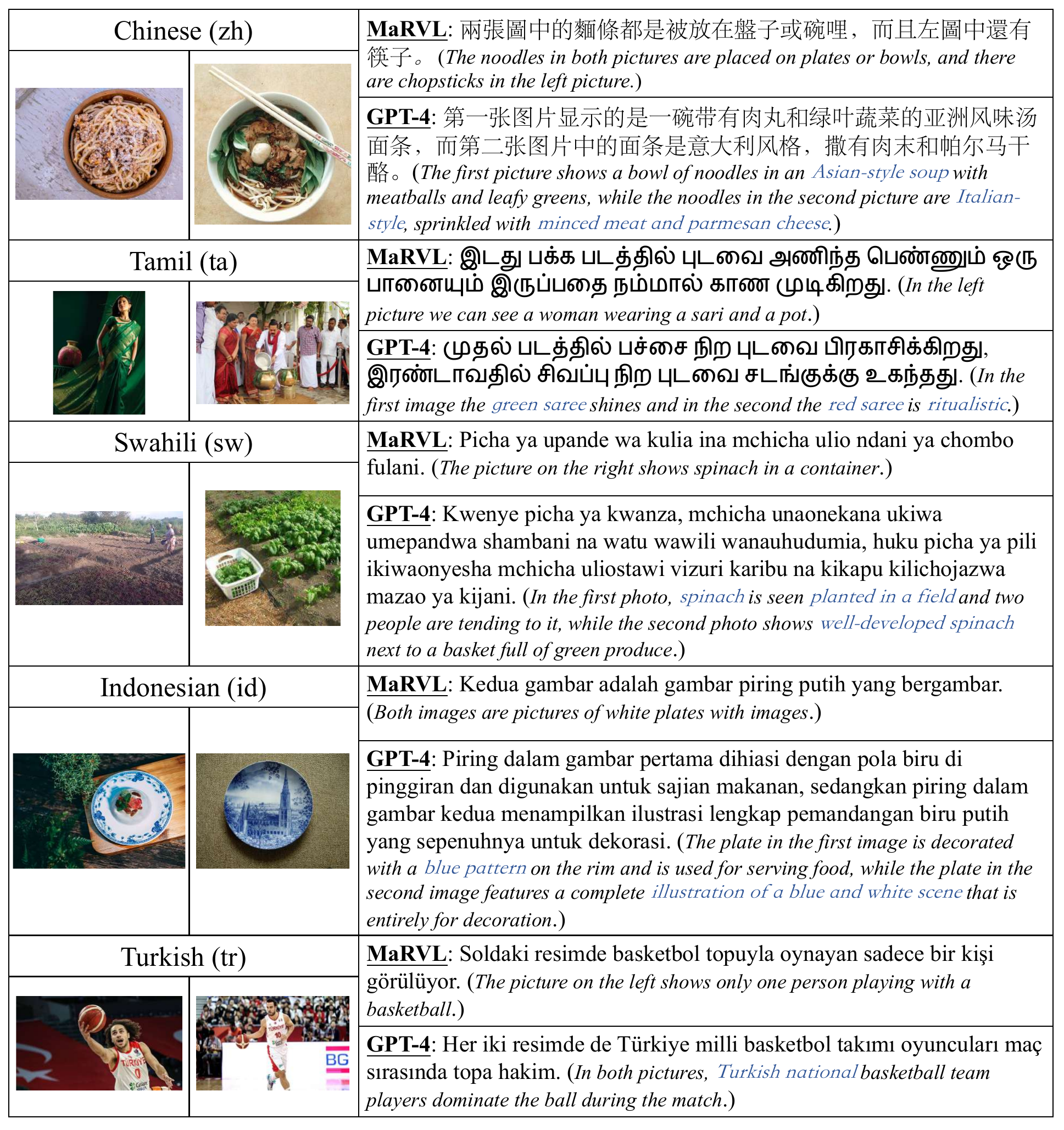}
    \caption{Additional case studies in the cultural pairwise caption generation task across languages, with cultural-relevant descriptions highlighted in blue.}
    \label{fig:more_case_study}
\end{figure*}

Here, we provide supplementary cases, as shown in Figure \ref{fig:more_case_study}. It is apparent that, across all languages, the descriptions generated by GPT-4 adeptly capture cultural concepts with precision. Moreover, these descriptions demonstrate a heightened level of granularity in terms of cultural relevance, thereby substantiating the model's superior quality and enhanced cultural awareness capabilities.

\begin{table}[t]
\centering
\resizebox{0.49\textwidth}{!}{
\begin{tabular}{l|cccccc}
\toprule
Task & ZH   & TA   & SW   & ID   & TR & Total \\ \midrule
CC & 1012 & 1242 & 1108 & 1128 & 1180 & 5670 \\
PC & 94 & 83 & 78 & 95 & 79 & 429 \\ 
CTS  & 188 & 166 & 156 & 190 & 158  & 858 \\
NE & 25 & 15 & 18 & 21 & 29 & 108 \\
\bottomrule
\end{tabular}}
\caption{Statistics pertaining to our probing data samples across the three tasks, including Culture checking (CC), Pairwise Captioning (PC), and Culture Tag Selection (CTS) along with the number of evaluators (NE) sourced from the crowdsourcing platform. \label{tb:eval-data}}
\end{table}

\CJK{UTF8}{gbsn}
\begin{table*}
\centering
\begin{tabularx}{2\columnwidth}{X}
\toprule
\multicolumn{1}{c}{\small Culture Checking} \\
\small \textbf{Prompt:}

任务：我会给你两张图片和一个描述句子，你需要判断这个描述句子是不是正确描述两张图片中的{\#concept}（关于{\#concept}的描述正确即可）。如果给定的句子描述内容为正确的，则输出为[真]，否则输出为[假]。 

\textit{Task: I will provide you with two images and a descriptive sentence. Your task is to determine whether the descriptive sentence accurately describes the concept ({\#concept}) depicted in the two images. If the given sentence correctly describes the content related to {\#concept}, the output should be [True]; otherwise, the output should be [False].}

描述句子:{\#Caption}  

\textit{Sentence: {\#Caption}.}

确保严格遵循以下输出格式。 

\textit{Ensure strict adherence to the following output format.}

[输出格式]:  

\textit{[Output Format]:}

答案: [真/假]。 

\textit{Answer: [True/False].}

解释: [解释你给出答案的理由]。 

\textit{Explanation: [Explain the reasoning behind your answer].}

\small \textbf{Output}

根据提供的图片，左图显示的是一轮满月而不是月饼，右图展示的是一个月饼，与描述中的悬挂的灯笼不符。因此，描述句子不正确描述了这两张图片。 

\textit{According to the provided images, the left picture depicts a full moon, not a mooncake, while the right picture shows a mooncake. This is inconsistent with the description of hanging lanterns. Therefore, the descriptive sentence does not accurately describe the content in these two images.} \\  \midrule

\multicolumn{1}{c}{\small Pairwise image captioning} \\
\small \textbf{Prompt:}

任务：我会给你两张图片，你需要生成一个句子，能够同时描述这两张图片中的{\#concept} (你的描述重点将是关于{\#concept})。文本内容长度大约10-40个字。 

\textit{Task: I will provide you with two images, and you need to generate a sentence that can simultaneously describe the {\#concept} in both pictures (with the focus on {\#concept}). The text should be approximately 10-40 words in length. }

确保严格遵循以下输出格式。 

\textit{Ensure strict adherence to the following output format.}

[输出格式]:  

\textit{[Output Format]:}

描述句子:  

\textit{Sentence:}

[描述内容] 

\textit{[Description]}

\small \textbf{Output}
第一张照片展示了室内浪漫的七夕布置，而第二张则是户外放天灯的庆祝活动。 

\textit{First image showcases romantically decorated indoor setting for Qixi Festival, while the second one captures an outdoor celebration with sky lanterns.} \\  \midrule

\multicolumn{1}{c}{\small Culture Tag Selection} \\
\small \textbf{Prompt:}

Task: Given two images, select the most appropriate cultural tag from the provided options. Options: Indonesian, Mandarin Chinese, Swahili, Tamil, and Turkish.",

Ensure strict adherence to the specified output format.

Output format:

Answer: [The option you chose].

\small \textbf{Output}
Mandarin Chinese \\ 
\bottomrule
\end{tabularx}
\caption{Example instances of our probing methodology encompass three tasks in Chinese and English prompts, please refer to our code for other languages. The italicized text has been translated from Chinese to English using Google Translate for a better understanding.}
\label{tb:probing_cases}
\end{table*}

\end{document}